\def\year{2021}\relax
\documentclass[letterpaper]{article} 
\usepackage{aaai21}  
\usepackage{times}  
\usepackage{helvet} 
\usepackage{courier}  
\usepackage[hyphens]{url}  
\usepackage{graphicx} 
\usepackage{natbib}  
\usepackage{caption} 
\usepackage{amsmath}
\usepackage{amssymb}
\usepackage{adjustbox}
\usepackage{subfigure}

\title{Sparse Depth Completion with Semantic Mesh Deformation Optimization}

\author {
    Bing Zhou, Matias Aiskovich, Sinem Guven\\
}
\affiliations {
    IBM Research\\
    \{bing.zhou, matias.aiskovich, sguven\}@ibm.com
}

\begin{document}

\maketitle

\begin{abstract}
Sparse depth measurements are widely available in many applications such as augmented reality, visual inertial odometry and robots equipped with low cost depth sensors. Although such sparse depth samples work well for certain applications like motion tracking, a complete depth map is usually preferred for broader applications, such as 3D object recognition, 3D reconstruction and autonomous driving. Despite the recent advancements in depth prediction from single RGB images with deeper neural networks, the existing approaches do not yield reliable results for practical use. In this work, we propose a neural network with post-optimization, which takes an RGB image and sparse depth samples as input and predicts the complete depth map. We make three major contributions to advance the state-of-the-art: an improved backbone network architecture named \textit{EDNet}, a semantic edge-weighted loss function and a semantic mesh deformation optimization method. Our evaluation results outperform the existing work consistently on both indoor and outdoor datasets, and it significantly  reduces the mean average error by up to 19.5\% under the same settings of 200 sparse samples on NYU-Depth-V2 dataset.

\end{abstract}

\section{Introduction}
Depth sensing is one of the key steps to understanding our world in 3D, which is important in a variety of applications, such as augmented reality, human-computer interaction, autonomous driving and 3D object model reconstruction~\cite{du2020depthlab,ren2011depth,wang2019pseudo,izadi2011kinectfusion}. Existing depth sensors, such as LiDARs, infrared-based depth cameras or stereo cameras are not yet widely used due to their limitations~\cite{hecht2018lidar,huang2018joint}. LiDARs usually have high costs, which make them typically only available in high end vehicles, while infrared-based depth camera accuracy is subject to lighting conditions~\cite{hecht2018lidar}. Stereo camera systems require careful calibration and the performance is subject to the textures of the target surface~\cite{huang2018joint}. 
Thus, depth prediction from single camera provides a more favorable and practical way to enable 3D sensing capabilities for many applications.

\begin{figure}
    \centering
    \includegraphics[width=2.8in]{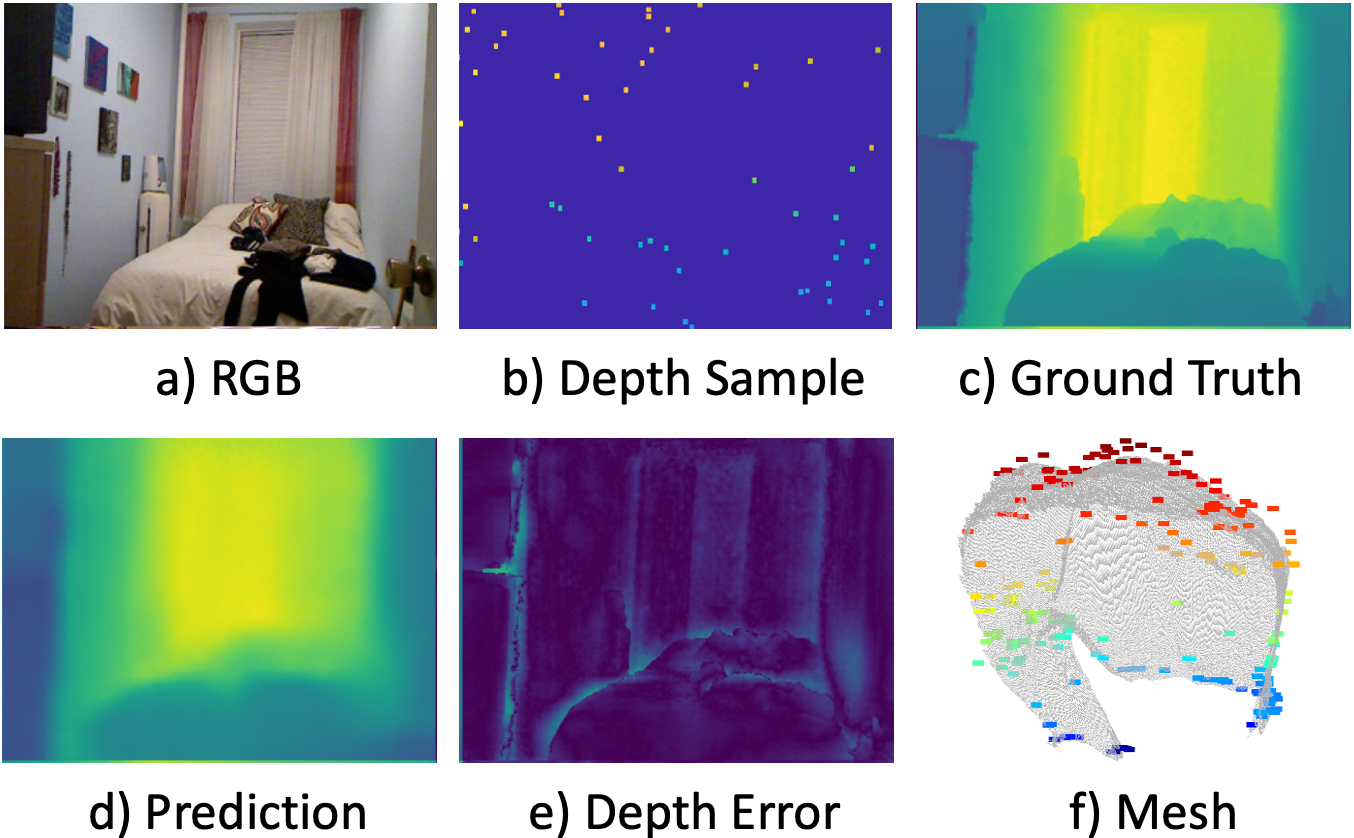}
    \caption{Depth completion sample result: a) and b) are RGB image and sparse depth samples used as input; c) and d) are the depth map ground truth and predicted result from~\cite{mal2018sparse}; e) shows the prediction error; f) shows the sparse depth samples in 3D and a reconstructed mesh.}
\vspace{-5pt}
    \label{fig:teaser}
\end{figure}


Despite the recent advancements in  depth  prediction  from  single  RGB  images  with  deep neural networks ~\cite{laina2016deeper,eigen2015predicting}, existing approaches still have limited accuracy and reliability. 
For example, Laina \emph{et al.}~\cite{laina2016deeper} achieved an average root mean error of about a half meter in indoor scenarios on the NYU-Depth-V2 dataset~\cite{silberman2012indoor}. However, such accuracy is still not sufficient for many applications.
Since sparse depth samples are readily available in many applications, such as augmented reality and Simultaneous Localization and Mapping (SLAM) applications~\cite{mur2017orb}, they can be potentially used to improve the depth estimation robustness and accuracy.

It seems straightforward to extend existing networks trained for RGB-to-depth prediction by providing them an extra sparse depth channel as input. Following this route, Ma~\textit{et al.}~\cite{mal2018sparse} proposed to train a Convolutional Neural Network (CNN) to learn a deep regression model for depth map prediction, which significantly improved the accuracy over existing RGB-based approaches~\cite{roy2016monocular,eigen2015predicting,laina2016deeper} and achieved the best accuracy over other RGB and depth fusion approach~\cite{liao2017parse}. 
Although significant improvement is observed compared to the RGB-based approach, directly feeding sparse depth samples into CNNs has some potential problems: it is difficult to learn effective information from sparse depth samples using the CNN,  which results in generally copying and interpolating the input depth samples. 

Figure~\ref{fig:teaser} shows an example of an RGB image from NYU-Depth-V2 dataset and prediction result using the method in Ma~\textit{et al.}~\cite{mal2018sparse}. Figure~\ref{fig:teaser}(b) shows the extra depth channel as input which contains 200 depth samples. Figure~\ref{fig:teaser}(c) and Figure~\ref{fig:teaser}(d) are the ground truth and predicted depth map, from which we can see the predicted depth map captures the overall depth distribution relatively well. However, it appears blurry compared to the ground truth since it's not able to capture the sharp depth transitions between objects.
Figure~\ref{fig:teaser}(e) shows the difference between the ground truth and the predicted result, and we can observe that most major depth errors occur around the transition edges between adjacent objects, where sharp depth changes usually happen. Figure~\ref{fig:teaser}(f) shows the known sparse depth samples projected in 3D, and a mesh surface reconstructed by connecting the predicted depth samples projected in the same 3D coordinate space. There is a clear discrepancy between the predicted result and the ``ground truth'' sparse depth samples. This indicates that even with the sophisticated CNNs, the results are still not optimal, leaving a significant room for improvement. 




Based on our observations, to further advance the state-of-the-art in sparse depth completion, we propose two improvements: \textit{1) Reducing major errors on object boundaries as shown in Figure~\ref{fig:teaser}(e)}; and \textit{2) Further calibrating the predicted depth map from neural networks using sparse depth samples as shown in Figure~\ref{fig:teaser}(f)}. 
In order to achieve the above improvements, our work provides three contributions. First, we design a more sophisticated neural network named \emph{EDNet}, which outperforms the existing state-of-the-art approach proposed in Ma~\textit{et al.}~\cite{mal2018sparse} on the same datasets. Next, we train the model with our customized semantic edge-weighted loss function, which further improves accuracy. Last, we prove that results from neural networks are sub-optimal, and propose a post-optimization approach based on semantic mesh deformation leveraging known sparse depth samples, which greatly improves the depth map accuracy both quantitatively and qualitatively. 
\emph{EDNet} predicts better local 3D shapes of the objects from the input RGB image, and the post deformation optimization calibrates such local shapes to more accurate absolute locations in 3D space. The combination of the probabilistic neural network prediction and deterministic post optimization brings better local and global accuracy, resulting in much improved results.

\begin{figure*}[!ht]
    \centering
    \includegraphics[width=6.3in]{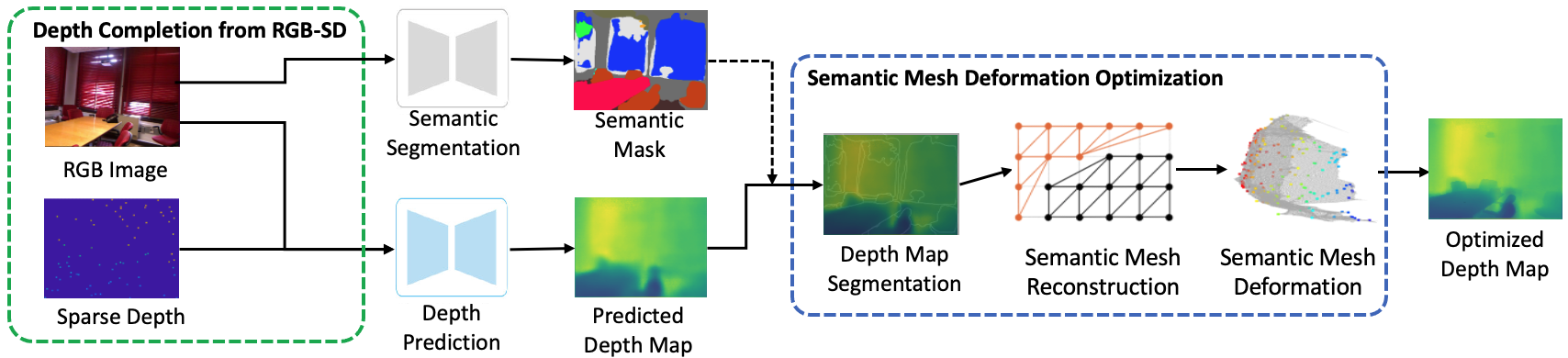}
    \caption{We predict a depth map from sparse depth samples and single RGB image using a neural network trained with semantic edge-weighted loss, and further optimize the result with semantic mesh deformation optimization.}
\vspace{-5pt}
    \label{fig:design}
\end{figure*}
\section{Related Work}
\textbf{RGB-based Depth Prediction.}
Depth prediction from RGB images has been an active research area due to the need for depth sensing in many applications where only RGB cameras are available. Several researchers estimate depth from RGB images using hand-crafted features, which  inevitably introduce noise~\cite{karsch2012depth,konrad20122d}. With the rapid progress in deep learning, neural networks achieved a lot of success in depth prediction~\cite{eigen2014depth,mousavian2016joint,xu2017multi}. 
 Eigen~\textit{et al.}~\cite{eigen2014depth} adapt a two-scale CNN to estimate depth from a single RGB image with  a scale invariant error loss function to improve both global and local accuracy. Laina~\textit{et al.}~\cite{laina2016deeper} propose a fully convolutional residual network for depth prediction, which outperforms 
 the results in~\cite{eigen2014depth}. Mousavian \textit{et al.}~\cite{mousavian2016joint} jointly perform semantic segmentation and depth estimation with deep convolutional networks, which significantly reduced the prediction error. Xu~\textit{et al.}~\cite{xu2017multi} propose a deep model that fuses complementary information derived from multiple CNN outputs to further improve the result. Although the accuracy is much improved with these efforts, the depth map predicted from RGB image only is still not accurate enough for practical use.



\textbf{Depth Completion from Sparse Depth Samples.}
To further improve the accuracy and robustness of depth prediction algorithms, researchers are seeking improvements using sparse depth samples, which already exist in many applications.
Yuki~\textit{et al.}~\cite{Tsuji2018NonGuidedDC} propose an adversarial network based on sparse depth inputs for depth completion to optimize the depth maps.
Xu~\textit{et al.}~\cite{xu2019depth} combine a prediction network with a specific diffusion refined network to regularize the constraints between depth and normal for depth completion.
Liao~\textit{et al.}~\cite{liao2017parse} fuse the depth samples from a 2D laser scanner with RGB images for depth prediction, which outperforms the RGB-only approaches. Cheng~\textit{et al.}~\cite{xinjing2018spatial} 
and Park~\textit{et al.}~\cite{nonlocal2020} 
both obtain great results on the NYU-Depth-v2 dataset,  but they require far more depth samples as input compared to others. Ma~\textit{et al.}~\cite{mal2018sparse}  introduce a single deep regression network to learn directly from the RGB-D raw data, which learns a better cross-modality representation for RGB and sparse depth, and achieves the best accuracy to date. Even though significant improvements are achieved from the more sophisticated neural networks, we find there is still a large room for improvement in the results from such existing work. 

\section{Method}

Figure~\ref{fig:design} shows the overall architecture of our approach, which consists of two major components: \emph{Depth Completion from RGB-SD} and \emph{Semantic Mesh Deformation Optimization}. The first component predicts the depth map with our proposed network, which captures the overall shapes of objects. The second component further calibrates the absolute errors at all pixels with semantic mesh deformation, enhancing the overall accuracy and appearance.

\subsection{Depth Completion from RGB-SD}
We propose an \textit{E}fficient \textit{D}epth \textit{Net}work -- \emph{EDNet} -- a fully convolutional encoder-decoder neural network architecture to predict the completed depth map. Additionally, we also propose a semantic edge-weighted loss function to improve the model training. 

\subsubsection{EDNet Architecture.} 

\begin{figure}
    \centering
    \includegraphics[width=3.3in]{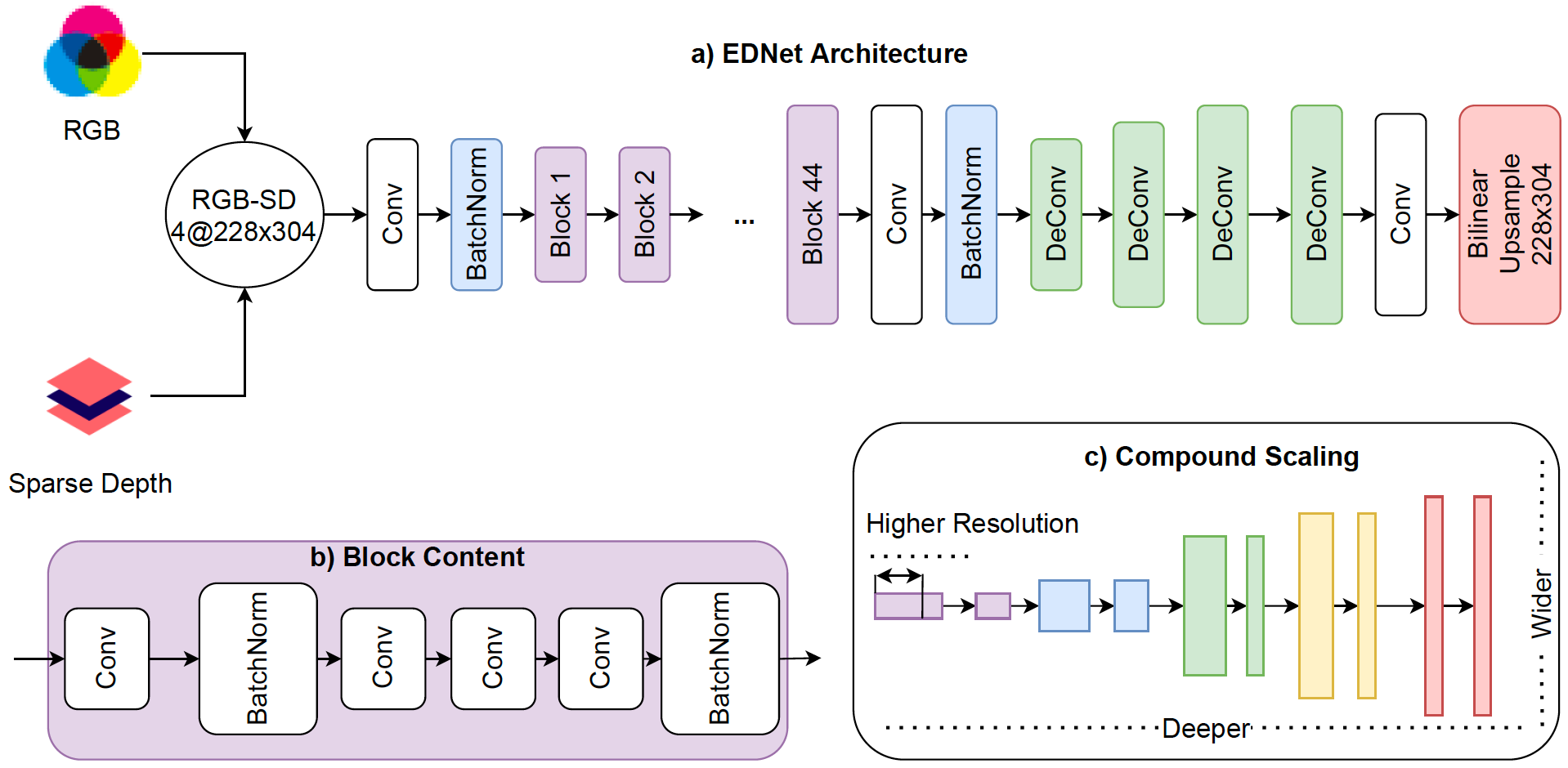}
    \caption{EDNet architecture and compound scaling blocks.}
    \label{fig:efficinet}
\end{figure}

We first adopt a general fully convolutional encoder-decoder architecture for our problem as it has shown best performance in multiple existing work~\cite{mal2018sparse, laina2016deeper}. Based on such high level design architecture, we build our customized encoder layer blocks and up-sampling layers.
Figure~\ref{fig:efficinet}(a) shows the overall structure of EDNet.

The encoder layers take the 4 channel RGB-SD data as input, followed by a convolutional layer and batch normalization layer. We also design a series of building blocks as shown in Figure~\ref{fig:efficinet}(b) as the encoding layers.
Each block contains a convolution layer, a batch normalization layer, three convolution layers and followed by another batch normalization layer.
For each block, it adopts a compound scaling method, which consists of scaling all dimensions (width, depth, and resolution) with a constant ratio inspired from~\cite{tan2019efficientnet}, as shown in Figure~\ref{fig:efficinet}(c). Deeper networks with such compound scaling mechanism enables larger receptive field and more channels to capture more fine-grained features from larger input resolution, without bringing significant computation overhead, making deeper networks more efficient.
It is critical to balance all dimensions of network width, depth, and resolution during scaling in order to obtain better accuracy and efficiency. We follow the same compound scaling method in~\cite{tan2019efficientnet} and set \(d = \alpha^\phi\), \(w = \beta^\phi\) and \(r = \gamma^\phi\), where $\alpha$, $\beta$, $\gamma$ are constants for depth, width and resolution scaling factor. We use the suggested coefficients in~\cite{tan2019efficientnet},  \(\alpha = 1.2, \beta =1.1, \gamma = 1.15\), and scale up baseline network with different user-specified coefficient \(\phi\) that controls the resources available
for model scaling.



The decoding layers are composed of 4 upsampling layers, followed by a bilinear upsampling layer. We choose deconvolution with a 3x3 kernel size as the upsampling layers, which slightly outperforms the UpProj module used in Laina~\emph{et al.}~\cite{laina2016deeper}.
We tailor the model with input data of different modalities, sizes, and dimensions to apply to our problem. 
We vary the amount and sizes of blocks in our design to generate different versions of EDNet to accommodate different input data and trade-offs. In our implementation, we get the best performance with 44 blocks, which we name it as EDNet-44.

\textbf{Loss Function.}
Common choices of loss function for such regression problems are either based on absolute error ($\mathcal{L}_1$) or mean squared error ($\mathcal{L}_2$). $\mathcal{L}_2$ is more sensitive to outliers in the training process as it penalizes more heavily on large errors, which yields over-smoothed boundaries instead of sharp transitions~\cite{mal2018sparse}. Thus, we choose $\mathcal{L}_1$ as our default basic loss function for its simplicity and performance.
We denote the ground truth depth map as $D \in \mathbb{R}^{H \times W}$ and the predicted depth map as $D' \in \mathbb{R}^{H \times W}$, then the basic $\mathcal{L}_1$ is defined as:
\begin{equation}
    \mathcal{L}_1 = \frac{1}{N}\sum_{p\in P}\left|D(p)-D'(p)\right|
\end{equation}
where $p=(i,j)$ is the index of the pixel, $P$ is the index set of all pixels, $N$ is the number of pixels in the set $P$; $D(p)$ and $D'(p)$ are the values of the pixels in the ground truth and predicted depth maps.

\subsubsection{Semantic Edge-Weighted Loss Function.}\label{sec:smd}
With the basic $\mathcal{L}_1$ loss function, our model achieves slightly better results compared to the Ma~\emph{et al.}~\cite{mal2018sparse}, as shown in our evaluation (Section~\ref{sec:eva}). However, we also notice that major errors exist on the pixels of object boundaries.
Based on this observation, we propose a semantic edge-weighted loss function to further explore the opportunity for optimization. This requires two major steps: \textit{semantic edge extraction} and \textit{edge-weighted loss design}.


\textbf{Semantic Edge Extraction.}
First, we need to extract the semantic segmentation masks from the RGB image in order to isolate the individual object instances. We leverage an existing semantic segmentation network~\cite{zhou2018semantic} with pre-trained weights to generate the mask images. 
Next, we apply Canny edge detection~\cite{canny1986computational} on the mask images to extract the pixels on the edges of the object instances. Note that this is different from directly applying Canny edge detection on the input RGB image, which captures all the visual edges, rather than the \textit{semantic edges} that only contain the object boundaries. Visual edges in texture-rich objects should be ignored since they belong to the same object, which usually have depth continuity; while different objects usually have depth variations.

\textbf{Edge-Weighted Loss Design.} 
Let's assume the resulting edge image from Canny edge detection is represented as $E \in \mathbb{R}^{H \times W}$, where $E(p)=1$ means the pixel belongs to the edge and $E(p)=0$ means the pixel is within an object.
Since pixels on the edges produce larger errors, the most intuitive way is to apply different weights on such pixels when we calculate the $\mathcal{L}_1$ loss. We multiply the loss at such pixels with a constant value of $\alpha$, which determines the weight. Thus, the modified loss function is shown as follows:
{\small
\begin{equation}
    \mathcal{L} = \frac{1}{N}\sum_{p\in P}(1-E(p)) \cdot \left|D(p)-D'(p)\right| + 
    \alpha \cdot E(p) \cdot \left|D(p)-D'(p)\right|
\end{equation}
}

\subsection{Semantic Mesh Deformation Optimization}

\begin{figure*}
    \centering
    \includegraphics[width=5in]{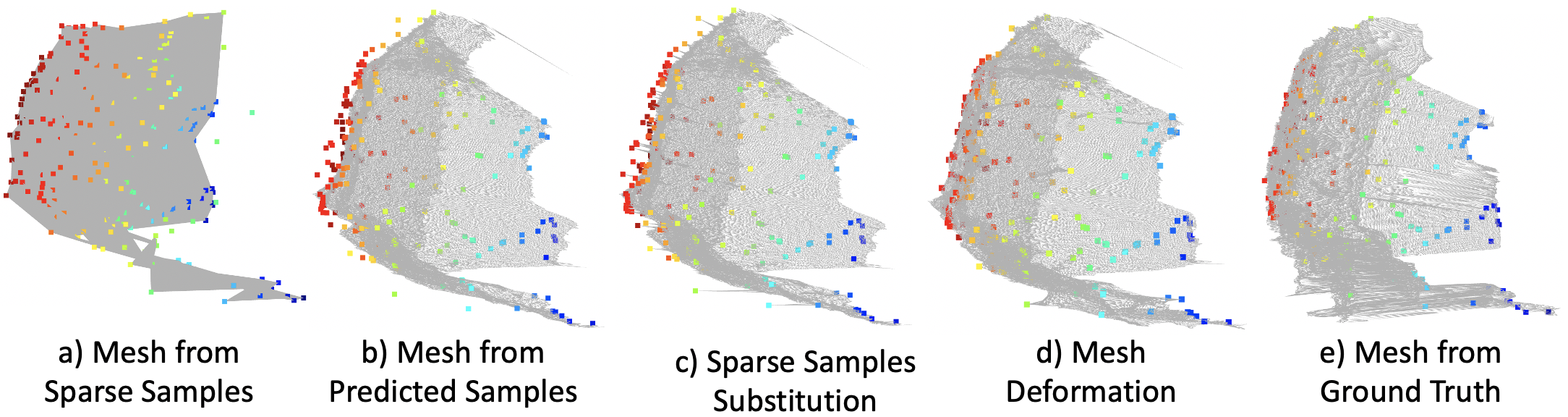}
    \caption{Visualization of depth map results, colored dots represent input sparse depth samples: a) mesh reconstructed from sparse samples; b-d) results from predicted depth map and optimizations; e) result based on ground truth depth map.}
\vspace{-7pt}
    \label{fig:mesh}
\end{figure*}


We also propose a semantic mesh deformation approach to optimize all the depth pixels, which consists of three major steps: \textit{semantic mesh reconstruction}, \textit{semantic mesh deformation} and \textit{depth map generation}.

\subsubsection{Semantic Mesh Reconstruction}
Semantic mesh reconstruction means that we reconstruct a mesh for each object instance in the RGB image so that we can calibrate such meshes independently rather than optimizing a single mesh reconstructed from the full depth map globally. Calibrating the meshes independently allows sharp transitions between adjacent objects at different depth distances. The detailed steps are described below:

\textit{Step 1: Semantic Depth Map Segmentation.}
The first step is to segment the predicted depth map $\mathcal{D'}$ into a list of semantic depth maps $\mathcal{D'}_d = \{D'_1, D'_2, \dots, D'_N\}$ according to the semantic segmentation masks obtained in semantic edge-weighted loss computation, where each $D'_i$ consists of the portion of depth map of one object, $N$ is the total number of segmented objects. We also divide the sparse depth samples $D_s$ into groups based on the segmentation mask and associate them to each semantic depth map $D'_i$. We represent such grouped sparse samples as $\mathcal{S}=\{S_1,S_2,\dots,S_N\}$, where each $S_i$ contains the sparse samples falling into the corresponding mask.
We remove those grouped depth samples $D'_i$ and grouped sparse samples $S_i$ from $D'_d$ and $S$ if $S_i$ contains no sparse depth samples due to none of them falls into the associated object mask.

\textit{Step 2: Depth to 3D Projection.}
The next step is to project segmented depth maps to 3D space. Given the camera intrinsic parameters, we can calculate the 3D coordinates of each depth sample following the projection rules.
Thus, we convert the semantic depth maps $\mathcal{D'}_d = \{D'_1, D'_2, \dots, D'_N\}$ to a list of point clouds in 3D coordinates $\mathcal{P}=\{P_1, P_2, \dots, P_N\}$, where each point cloud $P_i$ consists of the predicted depth samples of an object instance.



\begin{figure}
    \centering
    \includegraphics[width=2.5in]{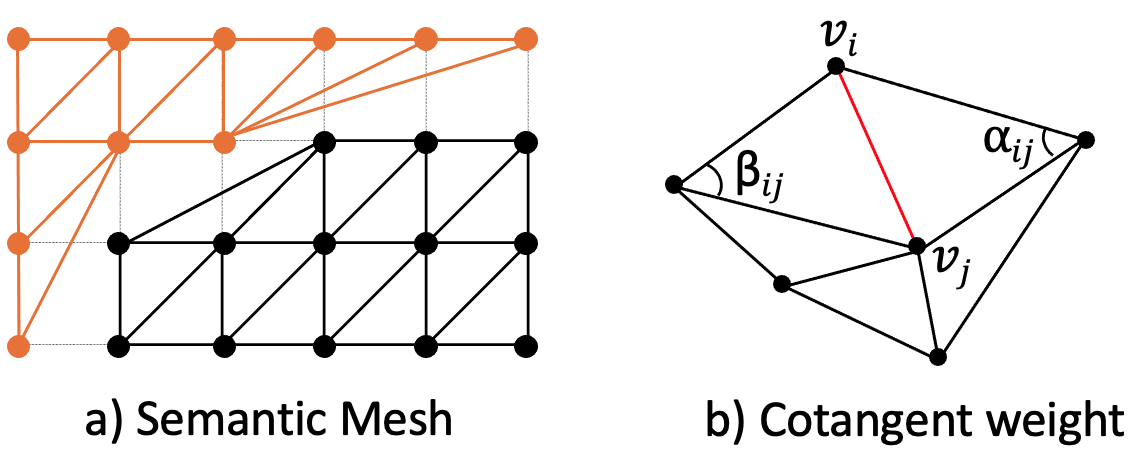}
    \caption{Semantic mesh and cotangent weight illustration.}
\vspace{-8pt}
    \label{fig:connection}
\end{figure}

\textit{Step 3: Semantic Mesh Reconstruction.}
After the point cloud list $\mathcal{P}$ is obtained, we reconstruct a mesh for each point cloud $P_i$ so that we can further optimize it. Note that we can not create meshes using common methods, such as Poisson surface reconstruction~\cite{kazhdan2006poisson}, since our goal is to calibrate the input vertex positions, and not to generate a smooth surface with augmented vertices. Instead, we simply create a mesh by assigning edge connections between nearby vertices. We leverage Delaunay triangulation algorithms~\cite{lee1980two} to create triangular connections for mesh generation. Figure~\ref{fig:connection}(a) shows an illustration of two semantic meshes created by connecting the vertices within each point cloud group. 
We denote the reconstructed meshes for the object instances as $\mathcal{M}=\{M_1,M_2,\dots,M_N\}$. The next step is to optimize the vertex positions of each $M_i$ via semantic mesh deformation leveraging the corresponding sparse depth samples $S_i$.

\subsubsection{Semantic Mesh Deformation}
The reconstructed object instance meshes $\mathcal{M}=\{M_1,M_2,\dots,M_N\}$ captures the overall shape for each object viewed from the camera. And each mesh $M_i$ has an associated group of sparse samples $S_i$, which we call anchors. The goal of this mesh deformation is to apply transformation and adjustments on the mesh $M_i$ to make the corresponding vertices to be as close as possible to the anchors while minimizing the overall shape changes of the mesh. We choose to apply an ``as rigid as possible'' mesh deformation approach~\cite{sorkine2007rigid} to retain the overall shapes predicted from the neural network, while calibrating the mesh to correct position.

For each mesh $M_i$ and the corresponding sparse depth sample $S_i$, we deform the mesh to such ``anchor'' points by minimizing the energy function:
\begin{equation}
 E=\sum_{i} \sum_{j \in \mathcal{N}(i)} w_{i j}\left\|\left(\mathbf{p}_{i}^{\prime}-\mathbf{p}_{j}^{\prime}\right)-\mathbf{R}_{i}\left(\mathbf{p}_{i}-\mathbf{p}_{j}\right)\right\|^{2}   
\end{equation}
where $\mathbf{R}_{i}$ are the rotation matrices that we want to optimize, and $p_i$ and $p_i^{'}$ are the vertex positions before and after the optimizations, i.e., $p'_i\in S_i$ and $p_i$ is the corresponding vertex in $M_i$. $\mathcal{N}(i)$ is the set of neighbors of vertex $i$. The weights $w_{i j}=\frac{1}{2}(cot\alpha_{ij} + cot\beta_{ij})$ are cotangent weights as shown in Figure~\ref{fig:connection}(b). Since the cotangent weights are representations based on angles, they preserve angles and shapes better during the deformation. We then optimize this energy function in an iterative way to get the rotation matrices $\mathbf{R}_{i}$, based on which we deform the semantic mesh $M_i$ by applying the rotation matrices to the associated vertices.




An example of the deformed mesh is shown in Figure~\ref{fig:mesh}. Figure~\ref{fig:mesh}(a) shows a mesh reconstructed from the available sparse depth samples (i.e., anchors) only, which is very coarse-grained. The anchors are shown as colored dots. Figure~\ref{fig:mesh}(b) shows the reconstructed mesh based on the predicted depth map. It captures much more fine-grained shapes, however it also shows obvious errors when matching to the known anchors. A naive way to optimize it is by substituting corresponding vertices in the mesh with the known anchors (shown in Figure~\ref{fig:mesh}(c)). Figure~\ref{fig:mesh}(d) shows the mesh after deformation to these anchors and Figure~\ref{fig:mesh}(e) is the mesh reconstructed from the ground truth depth map. We can observe that mesh deformation brings overall improvements to all the pixels, resulting in the closest result to Figure~\ref{fig:mesh}(e) - the ground truth.


\subsubsection{Depth Map Regeneration}
To generate the final depth map from the deformed meshes, we project the vertices of each mesh from 3D to 2D from the camera view, and calculate the distances to the camera. For those meshes without supporting anchors, we simply keep them unchanged and assign the original depth values for the corresponding pixels.

\section{Experiments}
We implement the neural networks using PyTorch and train the model on a computing cluster equipped with Nvidia V100 GPUs. 
For most experiments, we use a batch size of 16 and train for 15 epochs with a learning rate starting at 0.01, and reduced by 20\% every 5 epochs.

\textbf{Dataset.} 
We validate our approach on the NYU-Depth-v2 dataset~\cite{silberman2012indoor}, which consists of RGB and depth images collected from indoor scenes.
For training, we sample spatially evenly 
from the training dataset, generating roughly 48k synchronized RGB-depth image pairs, following the same setup in~\cite{mal2018sparse}. 
For fair comparison with existing work~\cite{mal2018sparse,laina2016deeper, eigen2014depth}, we resize the original images of size 640x480 to half, then center cropped to a size of 304x228 and test the final performance with the 654 images in test set.

We also use the KITTI dataset~\cite{Geiger2012CVPR} for outdoor environment evaluation, which includes both camera and LiDAR measurements. For fair comparison, we follow the same procedure as Ma~\emph{et al.}~\cite{mal2018sparse} and use all 46000 images from the training, and a random subset of 3200 images for the final evaluation. We use the bottom half of the images with a resolution of 921x228 as input, since the LiDAR has no measurement to the upper part of the images.

\textbf{Data Augmentation.}
We augment the training data during the training process through a series of transformations.
RGB images are first scaled by a random number $s \in [1, 1.5]$ and a random rotation. Then we apply color jitter to change the brightness, contrast, and saturation of color images. Finally, the images are flipped with a 50\% chance.
Apart from augmenting RGB images, we also need to augment sparse depth samples and semantic edges. 
 To augment the depth channel, we apply the same scaling transform by dividing the depth values by $s$ as it's opposite to the RGB. Then we apply the same rotation and flips as RGB image.
We also apply the same scaling, rotation, and flips transformations on the semantic edges. 

\textbf{Evaluation Metrics.}
We choose a set of metrics to evaluate our approach: root mean squared error (RMSE), mean absolute error (MAE), mean absolute relative error (REL) and $\delta_i$: percentage of predicted pixels where the relative error is within a threshold:
\begin{equation}
    \delta_{i}=\frac{\operatorname{card}\left(\left\{\hat{y}_{i}: \max \left\{\frac{\hat{y}_{i}}{y_{i}}, \frac{y_{i}}{\hat{y}_{i}}\right\}<1.25^{i}\right\}\right)}{\operatorname{card}\left(\left\{y_{i}\right\}\right)}
\end{equation}
where $y_i$ and $\hat{y}_i$ are the ground truth and the prediction, and card is the cardinality of a set. A higher $\delta_{i}$ means higher accuracy. In addition, we also plot the cumulative distribution function (CDF) curve to show the detailed error distributions.
\section{Results}\label{sec:eva}

\subsection{Indoor Dataset Results}
We compare our results with the existing RGB based methods~\cite{roy2016monocular,eigen2015predicting,laina2016deeper} and RGB-D based methods~\cite{liao2017parse,mal2018sparse,deepcorrelation2019}. To achieve the best result in our design, we use EDNet-44 trained with semantic edge-weighted loss function, and further optimize the results with semantic mesh deformation. The quantitative results are listed in Table~\ref{tab:accuracy}. 

The results show all RGB-D methods outperform the RGB based methods. Even with only 20 sparse depth sample, the overall accuracy is greatly improved. Our approach consistently outperforms the state-of-the-art~\cite{mal2018sparse} and others~\cite{liao2017parse, deepcorrelation2019} with different amount of sparse samples. With 200 sparse samples, which is common in most applications such as AR or SLAM, our results are significantly better compared to Ma~\emph{et al.}~\cite{mal2018sparse} in all the metrics.
Next, we will evaluate each component in our design and demonstrate how each component helps to improve the accuracy.

\begin{table}
\begin{adjustbox}{width=1.0\columnwidth,center}
\begin{tabular}{|l|c|c|c c|c c c|}
\hline
Input & $\#$Samples & Method & RMSE & REL & $\delta_1$ & $\delta_2$ & $\delta_3$ \\
\hline\hline
RGB & 0 & Roy \textit{et al.}\cite{roy2016monocular} & 0.744 & 0.187 & - & - & - \\
    & 0 & Eigen \textit{et al.}\cite{eigen2015predicting} & 0.641 & 0.158 & 76.9 & 95.0 & 98.8 \\
    & 0 & Laina \textit{et al.}\cite{laina2016deeper} & 0.573 & 0.127 & 81.1 & 95.3 & 98.8 \\
\hline\hline
RGB-D & 225 & Liao \textit{et al.}\cite{liao2017parse} & 0.442 & 0.104 & 87.8 & 96.4 & 98.9 \\
     & 20 & Ma~\textit{et al.}\cite{mal2018sparse} & 0.351 & 0.078 & 92.8 & 98.4 & 99.6 \\
     & 20 & Ours & 0.338 & 0.077 & 93.2 & 98.5 & 99.6  \\
     & 50 & Zhong~\textit{et al.}\cite{deepcorrelation2019} & 0.715 & - & 65.5 & 90.1 & 97.4 \\
     & 50 & Ma~\textit{et al.}\cite{mal2018sparse}  & 0.281 & 0.059 & 95.5 & 99.0 & 99.7 \\
     & 50 & Ours & 0.278 & 0.059 & 96.3 & 99.2 & 99.7  \\
     & 200 & Zhong~\textit{et al.}\cite{deepcorrelation2019} & 0.531 & - & 80.9 & 95.1 & 98.7 \\
     & 200 & Ma~\textit{et al.}\cite{mal2018sparse} & 0.230 & 0.044 & 97.1 & 99.4 & 99.8 \\
     & 200 & Ours & \textbf{0.211}  & \textbf{0.039}  & \textbf{97.3} & \textbf{99.5} & \textbf{99.9} \\
\hline
\end{tabular}
\end{adjustbox}
\caption{Results comparison with existing work on NYU-Depth-v2 dataset, respective results are cited from the original papers.}\label{tab:accuracy}
\end{table}

\subsection{Ablation Study}
\subsubsection{Neural Network Architecture}
We evaluate the neural network architectures used in our work with standard $\mathcal{L}_1$ loss function without additional optimizations. We run the experiments with a series of EDNets (EDNet-32, EDNet-38 and EDNet-44) and compare the results with Ma~\emph{et al.}~\cite{mal2018sparse}, which reported best results so far on the same dataset. 
The quantitative results are shown in Table~\ref{tab:network}. For results with 100 sparse samples, our EDNet outperforms Ma~\emph{et al.}~\cite{mal2018sparse} consistently with about 10\% reduced MAE. 
With 200 sparse depth samples, the difference among different network architectures becomes smaller, while our EDNets still have slightly better results.
Note that these results are with the model trained with standard loss function, in the following evaluation, we show that EDNet-44 achieves the best result when trained with our semantic edge-weighted loss function. 


\begin{table}
    \centering
    \begin{adjustbox}{width=0.7\columnwidth,center}
    \begin{tabular}{ |l|c|c|c|c|  }
     \hline
     $\#$Samples&  Models & MAE & RMSE & REL\\
     \hline\hline
     100 & Ma~\emph{et al.}~\cite{mal2018sparse} & 0.161 & 0.270 & 0.056\\
         & EDNet-32 & 0.147 & 0.257 & 0.055\\
         & EDNet-38 & 0.149 & 0.261 & 0.055\\
         & EDNet-44 & 0.145 & 0.263 & 0.055\\
    \hline
     200 & Ma~\emph{et al.}~\cite{mal2018sparse} & 0.136 & 0.240 & 0.047 \\
         & EDNet-32 &  0.131 & 0.232 & 0.048\\
         & EDNet-38 & 0.137 & 0.239 & 0.049\\
         & EDNet-44 &0.132 & 0.233 & 0.049\\
    \hline
    \end{tabular}
    \end{adjustbox}
    \caption{Results of different base models without optimization.}
    \label{tab:network}
\end{table}





\subsubsection{Semantic Edge-Weighted Loss Function}
\begin{figure}
	\centering
	\subfigure[Mean average error.] {
		\includegraphics[width=1.55in]{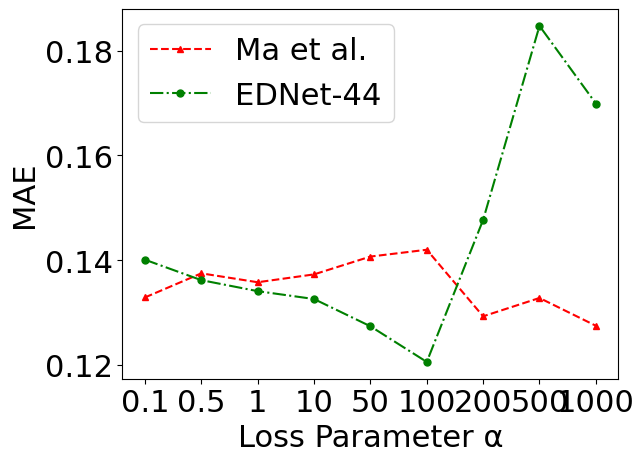}\label{fig_wo_da}
		}
	\subfigure[Root Mean Squared Error.]{
		\includegraphics[width=1.55in]{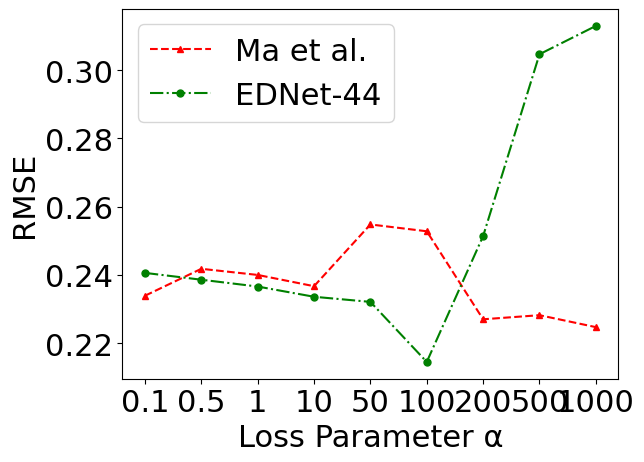}\label{fig_w_da}
	}
	\caption{Errors with different loss function parameter $\alpha$.} \label{fig:alpha}
\end{figure}

To evaluate the semantic edge-weighted loss function, we apply the loss function on both Ma~\emph{et al.}~\cite{mal2018sparse} and EDNet-44 and train the models with RGB images and 200 sparse samples. We test the parameter $\alpha$ with values between 0.1 to 1000, which indicates the weight we assign to edge pixels. $\alpha<1$ means smaller weights are assigned to the edges; $\alpha=1$ means all pixels are treated equally, i.e., no edge-weighted loss is applied; $\alpha>1$ means we put more weights on the edges. We show the MAE and RMSE errors of both models in Figure~\ref{fig:alpha}. We notice that as $\alpha$ is getting larger, EDNet-44 shows consistent improvements first, then the errors increase as $\alpha>100$. For Ma~\emph{et al.}~\cite{mal2018sparse}, the error curves are relatively stable when $\alpha<10$, showing little impact on the performance. Among all the results, EDNet-44 with $\alpha=100$ achieves the best result, 
a noticeable improvement compared to the result using standard $\mathcal{L}_1$ loss.


We also show the results of other EDNet versions with the semantic edge-weighted loss with $\alpha=100$, which yield the best result on EDNet-32 and EDNet-44. Table~\ref{tab:en_100} shows the results. Compared to the results in Table~\ref{tab:network} without edge-weighted loss of the same number of sparse samples, we get consistent improvements for all EDNet versions, demonstrating the effectiveness of our new loss function. 


\begin{table}
    \centering
    \begin{adjustbox}{width=0.6\columnwidth,center}
    \begin{tabular}{ |c|c|c|c|c|  }
     \hline
     Models & MAE & RMSE & REL\\
     \hline\hline
     EDNet-32 & 0.127 & 0.228 & 0.047\\
     EDNet-38 & 0.137  &  0.235 & 0.051\\
     EDNet-44 & 0.128  & 0.229 & 0.048\\
     \hline
    \end{tabular}
    \end{adjustbox}
    \caption{Results of different EDNet versions with the same loss function parameter $\alpha=100$.}
    \label{tab:en_100}
\end{table}

\subsubsection{Semantic Mesh Deformation Optimization}

\begin{figure*}[ht]
    \centering
    \includegraphics[width=5.0in]{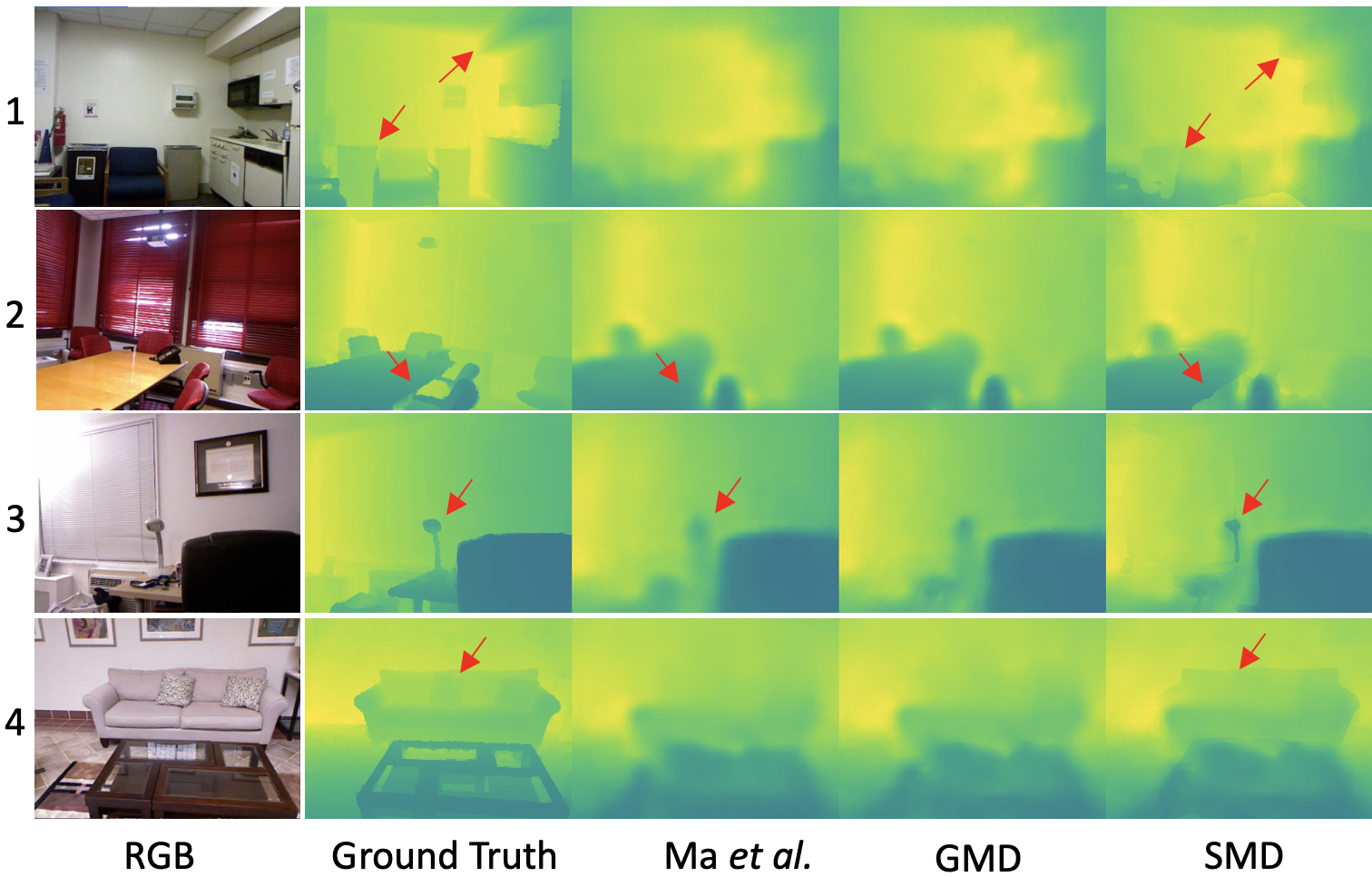}
    \caption{Sample results from NYU-Depth-V2 dataset comparison of Ma~\emph{et al.}~\cite{mal2018sparse} and our methods. Our SMD approach produces sharp transitions between adjacent objects which significantly increase the visual appearance.}
    \label{fig:smd_samples}
    \vspace{-5pt}
\end{figure*}

To evaluate the effectiveness of semantic mesh deformation optimization, we take the best prediction results from EDNet-44 trained with semantic edge-weighted loss with $\alpha=100$ as the baseline for optimization evaluation. We also compare the results of two optimization approaches:  \textit{Global Mesh Deformation (GMD)}, i.e., reconstruct a single mesh from the predicted depth map and deform it to all known sparse points; and \textit{Semantic Mesh Deformation (SMD)}, which is the chosen design in this paper.

\begin{figure}
    \centering
    \includegraphics[width=2.2in]{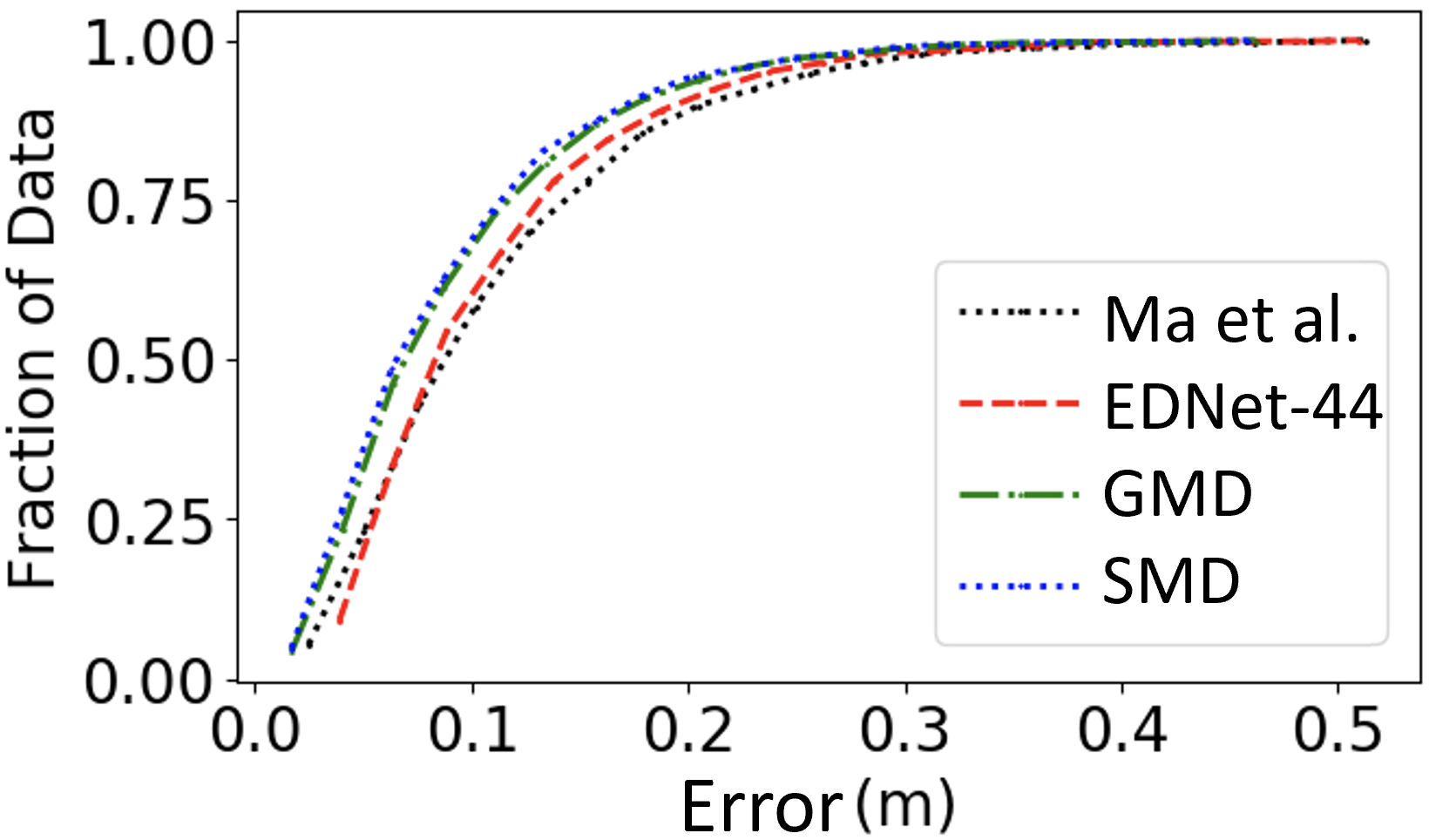}
    \caption{CDF of absolute errors of different approaches.}
    \label{fig:cdf}
\end{figure}

Figure~\ref{fig:cdf} shows the cumulative distribution function (CDF) curves of the absolute errors of Ma~\textit{et al.}~\cite{mal2018sparse}, EDNet-44 (EDNet without post-optimization), EDNet-44 with GMD and SMD optimizations. The CDF curve provides more details of the error distribution, which is a more comprehensive metric to evaluate the performance compared to MAE or RMSE. The X axis is the absolute error, and the Y axis is the percentage of samples having errors smaller than the error indicated in X axis.
In this figure, we can see all of our results outperform~\cite{mal2018sparse}. 
GMD shows considerable improvements due to the mesh deformation, and SMD  further increases this improvement by optimizing the object boundaries. Table~\ref{tab:compare} shows the quantitative results. Our SMD based approach outperforms~\cite{mal2018sparse} in all metrics, and it reduces the MAE error by 19.5\% (from 0.133 to 0.107), a significant improvement.

\begin{table}
    \centering
    \begin{adjustbox}{width=0.5\columnwidth,center}
    \begin{tabular}{ |c|c|c|c|c|  }
     \hline
     Models & MAE & RMSE & REL\\
     \hline\hline
     Ma~\textit{et al.} & 0.133 & 0.236 & 0.047\\
     EDNet-44 & 0.128  &  0.229 & 0.048\\
     GMD & 0.111  & 0.216 & 0.041\\
     SMD & \textbf{0.107}  & \textbf{0.211} & \textbf{0.039}\\
     \hline
    \end{tabular}
    \end{adjustbox}
    \caption{Results comparison of Ma~\textit{et al.} and our approach.}
    \vspace{-10pt}
    \label{tab:compare}
\end{table}

For qualitative comparison, Figure~\ref{fig:smd_samples} shows a few sample results of different approaches. Although GMD has proven better quantitative results than Ma~\emph{et at.}~\cite{mal2018sparse}, the visual appearances are similar. This is because GMD optimizes the depth map globally in an ``as rigid as possible'' way, thus the overall shape looks similar. However, we can still tell the difference (e.g., for the wall areas in the first sample, GMD result is slightly brighter than Ma~\emph{et at.}, more closer to ground truth). The SMD shows a noticeable improvement in terms of visual appearance with much sharper and cleaner object boundaries, as indicated by red arrows in Figure~\ref{fig:smd_samples}.
In contrast, such details are blurred out in Ma~\emph{et al.}~\cite{mal2018sparse} or GMD, which makes them difficult to recognize. This clearly shows the great improvement from SMD in qualitative aspect as well.

\subsection{Outdoor Dataset Results}
The KITTI outdoor dataset~\cite{Geiger2012CVPR} has a maximum distance of 100 meters, which makes it more challenging compared to an indoor dataset. We compare the results of our approach with existing approaches~\cite{eigen2014depth,mal2018sparse,liao2017parse}. Table~\ref{tab:kitti_outdoor} shows the results. Without sparse depth samples, existing approaches have a RMSE error larger than 6 meters, while our approach achieves the best accuracy with a RMSE around 4.5m. We also compare the results with different amounts of sparse points. Our results outperform Ma~\textit{et al.}~\cite{mal2018sparse} and others consistently, and reach the best accuracy of a RMSE of 3.1m with 500 sparse samples.

\begin{table}
\begin{adjustbox}{width=1.0\columnwidth,center}
\begin{tabular}{|l|c|c|c c|c c c|}
\hline
Input & $\#$Samples & Method & RMSE & REL & $\delta_1$ & $\delta_2$ & $\delta_3$ \\
\hline\hline
RGB & 0 & Eigen~\textit{et al.}\cite{eigen2014depth} & 7.156 & 0.190 & 69.2 & 89.9 & 96.7 \\
    & 0 & Ma~\textit{et al.}\cite{mal2018sparse} & 6.266 & 0.208 & 59.1 & 90.0 & 96.2 \\
    & 0 & Ours & \textbf{4.459} & \textbf{0.105} & \textbf{87.7} & \textbf{96.1} & \textbf{98.5} \\
\hline\hline
RGB-D & 225 & Liao \textit{et al.}\cite{liao2017parse} & 4.50 & 0.113 & 87.4 & 96.0 & 98.4 \\
     & 50 & Ma~\textit{et al.}\cite{mal2018sparse}  & 4.884 & 0.109 & 87.1 & 95.2 & 97.9 \\
     & 50 & Ours & \textbf{3.751} & \textbf{0.072} & \textbf{92.7} & \textbf{97.3} & \textbf{98.8}  \\
     & 100 & Ma~\textit{et al.}\cite{mal2018sparse} & 4.303 & 0.095 & 90.0 & 96.3 & 98.3 \\
     & 100 & Ours & \textbf{3.516}  & \textbf{0.064}  & \textbf{93.7} & \textbf{97.5} & \textbf{98.9} \\
     & 200 & Ma~\textit{et al.}\cite{mal2018sparse} & 3.851 & 0.083 & 91.9 & 97.0 & 98.6 \\
     & 200 & Ours & \textbf{3.344}  & \textbf{0.059}  & \textbf{94.4} & \textbf{97.6} & \textbf{98.9} \\
     & 500 & Ma~\textit{et al.}\cite{mal2018sparse} & 3.378 & 0.073 & 93.5 & 97.6 & 98.9 \\
     & 500 & Ours & \textbf{3.128}  & \textbf{0.054}  & \textbf{95.1} & \textbf{97.9} & \textbf{99.0} \\
\hline
\end{tabular}
\end{adjustbox}
\caption{Results comparison with existing work on KITTI dataset. 
}\label{tab:kitti_outdoor}
\end{table}

\section{Conclusion}

In this paper, we propose a new approach for sparse depth completion with three contributions: a sophisticated network architecture, a novel semantic edge-weighted loss function, and semantic mesh deformation optimization. 
It outperforms the existing work in all evaluation metrics on both indoor and outdoor datasets, and achieves a significant improvement of 19.5\% reduced MAE on NYU-Depth-V2 dataset under the same settings of 200 sparse samples.

{\small
\bibliography{egbib}
}

\end{document}